\documentclass[10pt,final,journal,twoside]{IEEEtran}

\usepackage{cite}
\usepackage{url}
\usepackage{graphicx}
\usepackage{booktabs}
\usepackage{tabularx}
\usepackage{multirow}
\usepackage{subfigure}
\usepackage{color}
\usepackage{subfigure}
\usepackage{epstopdf}
\usepackage[cmex10]{amsmath}
\usepackage{graphicx, subfigure}
\usepackage{amssymb}
\usepackage{stmaryrd}
\usepackage{comment}

\let\emptyset\varnothing

\usepackage{xspace}
\def\eg{\emph{e.g.}\xspace}

\begin{document}
\title{TagBook: A Semantic Video Representation\\ without Supervision for Event Detection}
\author{Masoud~Mazloom,~Xirong~Li*,~and~Cees~G.M.~Snoek,~\IEEEmembership{Senior~Member,~IEEE}
\thanks{
Copyright (c) 2013 IEEE. Personal use of this material is permitted. However, permission to use this material for any other purposes must be obtained from the IEEE by sending a request to pubs-permissions@ieee.org.

This research was supported by the STW STORY project, the Dutch national program COMMIT, the National Science Foundation of China (No. 61303184), the Fundamental Research Funds for the Central Universities and the Research Funds of Renmin University of China (No. 14XNLQ01), the Specialized Research Fund for the Doctoral Program of Higher Education (No. 20130004120006), and the Intelligence Advanced Research Projects Activity (IARPA) via Department of Interior National Business Center contract number D11PC20067. The U.S. Government is authorized to reproduce and distribute reprints for Governmental purposes notwithstanding any copyright annotation thereon. Disclaimer: The views and conclusions contained herein are those of the authors and should not be interpreted as necessarily representing the official policies or endorsements, either expressed or implied, of IARPA, DoI/NBC, or the U.S. Government.

*Corresponding author. 
M. Mazloom and C.G.M. Snoek are with the Informatics Institute, University of Amsterdam, Science Park 904, 1098 XH, Amsterdam, The Netherlands. C.G.M. Snoek is also with Qualcomm Research Netherlands. X. Li is with the Key Lab of Data Engineering and knowledge Engineering, School of Information, Renmin University of China, 100872 China. E-mail: xirong.li@gmail.com.}}
\markboth{IEEE TRANSACTIONS ON MULTIMEDIA,~Vol.~, No.~, April~2016}{Mazloom \MakeLowercase{\textit{et al.}}: TagBook: A
Semantic Video Representation without Supervision for Event Detection}

\maketitle
\begin{abstract}
We consider the problem of event detection in video for scenarios where only few, or even zero examples are available for training. For this challenging setting, the prevailing solutions in the literature rely on a semantic video representation obtained from thousands of pre-trained concept detectors. 
Different from existing work, we propose a new semantic video representation that is based on freely available social tagged videos only, without the need for training any intermediate concept detectors. We introduce a simple algorithm that propagates tags from a video's nearest neighbors, similar in spirit to the ones used for image retrieval, but redesign it for video event detection by including video source set refinement and varying the video tag assignment. We call our approach \emph{TagBook} and study its construction, descriptiveness and detection performance on the TRECVID 2013 and 2014 multimedia event detection datasets and the Columbia Consumer Video dataset. Despite its simple nature, the proposed TagBook video representation is remarkably effective for few-example and zero-example event detection, even outperforming very recent state-of-the-art alternatives building on supervised representations. 
\end{abstract}

\begin{IEEEkeywords}
Event Detection, Video tagging, Video search.
\end{IEEEkeywords}

\IEEEpeerreviewmaketitle

\section{Introduction}
\begin{figure}[!tb]
\centering
\includegraphics[width=\linewidth]{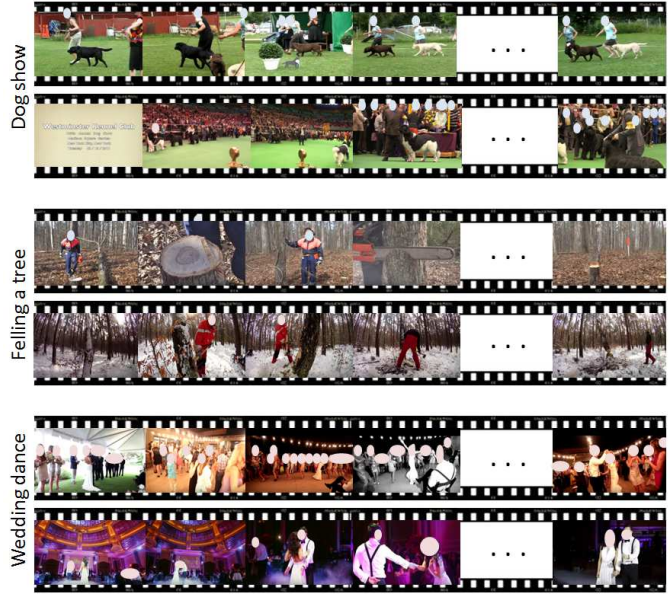}
\caption{Example videos for the events \emph{dog show}, \emph{felling a tree}, and \emph{wedding dance}. Despite the challenging diversity in visual appearance, each event maintains specific semantics in a consistent fashion. This paper studies whether an event representation based on tags assigned from the video's nearest neighbors can be an effective semantic representation for few- and zero-example event detection.} \label{fig:examples}
\end{figure}

\IEEEPARstart{T}{he} goal of this paper is to detect events such as \emph{dog show}, \emph{felling a tree}, and \emph{wedding dance} in arbitrary video content (Fig.~\ref{fig:examples}). The topic of event detection has a long tradition in the discipline of multimedia, see \cite{Ballan11,Lavee:understand,Mubarak:ijmir} for recent surveys. Early works considered knowledge-intensive approaches using relatively little video data, \eg~\cite{HAER00,BONZ01,LI01b, BABA02}. The state-of-the-art is to exploit big video data sets, such as the Columbia Consumer Video collection \cite{icmr11:CCV} and the TRECVID Multimedia event detection corpus \cite{MED13}, and to learn an event classifier from dozens of carefully labeled examples, \eg~\cite{Nakamasa:Tokyo,CVPR2012BBN,OneataICCV2013,xu2015discriminative}. However, as events become more specific, the harder it will be to find sufficient relevant examples for learning, even on the socially-tagged web \cite{HabibianICM2014,shih-fu:2014,WuCVPR2014}. Different from the dominant strategy in the event detection literature, we consider in this paper event detection scenarios where video examples of the event are scarce, and even completely absent.

The key to event detection is to have a discriminative video representation. Traditional video representations for event detection rely on low-level audiovisual features. Often combining bag-of-words derived from SIFT descriptors, MFCC audio features and space-time interest points~\cite{Jiang:Columbia,CVPR2012BBN,CVPR2012Kitware,McCloskey:MVA13,habibian:MVA13,super:icmr12,tmm2015-jiang-fast}, or by localizing temporal evidence by formulating the problem of video event detection as multiple instance learning in a low-level feature space~\cite{dynamicpooling:iccv, CVPR2014:dvmm, ECCV2014:dvmm}. Oneata~\emph{et al.} demonstrate the effectiveness and robustness of the improved dense trajectories with a Fisher vector encoding~\cite{OneataICCV2013,AXES}. Based on the great success in image recognition \cite{KrizhevskySH12}, learned representations derived from convolutional neural network layers are becoming popular for video event detection as well, e.g.~\cite{xu2015discriminative,MazloomICMR15,ye2015,bmvc2015-nagel}. Since most low-level representations have a high feature vector dimensionality, they reach good performance in the presence of sufficient positive training examples per event. However, the applicability of the low-level representation is limited when only a few positive event examples are available~\cite{ma:acm13,Mazloom:acm13,iccv2013-yang-few,mm2015-chang-few}. Especially in event detection scenarios where only a description of an event is available, i.e. without any video training example, the low-level representations by themselves are more or less useless. 
To tackle the scarcity of positive examples,
video samples which do not precisely describe an event but are still relevant to help detect the event are exploited in \cite{iccv2013-yang-few}.
A novel joint training protocol is developed in \cite{mm2015-chang-few} 
to simultaneously conduct event detection and recounting,
where the recounting model assists detection by filtering out noisy irrelevant information. 
We take a direction orthogonal to these works, 
aiming to find a semantic video representation capable of detecting events in the presence of few examples and even zero examples. 

Others have also studied semantic video representations in the context of few-example \cite{ma:acm13,HabibianICM2014,habibian14icmr,Mazloom:acm13} and zero-example \cite{shih-fu:2014,WuCVPR2014,habibian14icmr,jiang2014zero,chang2015semantic} event detection. All these works build a semantic representation on top of concept detectors such as `dog', `tree' and `groom'. Such an approach has become feasible to some extent thanks to the availability of thousand of concept annotations as part of the TRECVID benchmark~\cite{SMEA06,NAPH06} and the ImageNet challenge~\cite{ILSVRCarxiv14}, as well as social-tagged image and video resources~\cite{HabibianICM2014,shih-fu:2014,WuCVPR2014,mtap15-svetlana}. While this allows for few-example event detection indeed, the need for event examples has effectively been substituted for the even bigger problem of acquiring appropriate concept examples. Not to mention the computational demand for training the individual detectors. In contrast, we propose a new semantic video representation that is based only on freely available social tagged videos, without the need for training \emph{any} concept detectors. Before detailing our approach on how to arrive at the new video representation, we first discuss in more detail related work on semantic video representations.

\section{Related Work} \label{sec:related}
\subsection{Representations from supervised concepts}
There are good efforts for achieving a semantic representation by automatically recognizing concepts in a video's audiovisual content. The standard approaches attempt to train a classifier per concept and use the corresponding classifier confidence values as the building block for a video representation, which in turn is leveraged for event detection, e.g.~\cite{Shahram:Multi, Michele:SMV, HabibianCVIU14, Izadinia:2012, Mazloom:TMM2014, GKAL11, Sun_2014_CVPR, cvpr14:trc, MA13}. In~\cite{Shahram:Multi} for example, Ebadollahi \emph{et al.} employ 39 pre-defined concepts from the large scale concept ontology~\cite{NAPH06} for detecting events in broadcast news footage. Mazloom \emph{et al.} \cite{Mazloom:TMM2014} introduce a feature selection algorithm that learns the best concept representation for an event from a large bank of more than thousand concept detectors trained on ImageNet~\cite{ILSVRCarxiv14} and TRECVID~\cite{2013trecvidover}. Bhattacharya \emph{et al.} \cite{cvpr14:trc} leverage the temporal dynamics of concept detector scores in their representation using linear dynamical system models. Naturally, a semantic representation can be mixed with a low-level one, as successfully shown by Ma \emph{et al.} \cite{videoAtt:Haup}. All these works rely on carefully annotated images or video fragments to arrive at their concept detectors. Since it is hard to determine a priori what concepts will be needed, we prefer a more flexible video representation that builds its representation by learning from many weakly annotated web videos, e.g., YouTube videos with social tags.


\subsection{Representations from weakly-supervised concepts}
Weakly supervised web resources have been explored by others as well. In~\cite{HabibianICM2014}, Habibian \emph{et al.} harvest YouTube videos as a resource on which they base their representation. To accommodate for the ambiguity of the video descriptions they define a set of initial filters on the video collection, covering grammar and visualness of the descriptions, to assure the most reliable descriptions remain. To further alleviate the ambiguity an algorithm is proposed that learns an embedding of the joint video-description space. The embedding essentially groups several terms into topics to allow for a robust visual predictor, while maintaining descriptiveness. Rather than obtaining a semantic representation by training concepts over web video examples, Mazloom \emph{et al.} \cite{MazloomICMR15} propose an algorithm that learns a set of relevant frames as the concept prototypes, without the need for frame-level annotations.
Since the concept prototypes are a frame-level representation of concepts, they offer the ability of mapping each frame of a video into the concept prototype space, which can be leveraged for both few-example and zero-example event detection. Wu \emph{et al.}  \cite{WuCVPR2014} leverage off-the-shelf detectors as well as various video and image collections that come with textual descriptions to learn a large set of concept detectors using various multimedia features. 
To allow for zero-example detection, both the event description and concept detectors are mapped into the same textual space, in which their similarity is computed using the cosine distance. 
Chen \emph{et al.} \cite{shih-fu:2014} also start from a set of events and their textual descriptions. They first extract tags deemed relevant for the events. After verifying that the tags are meaningful and visually detectable, each tag is used as query on a photo sharing website. By doing so the authors harvest 400,000 image examples to build a representation containing a total of 2,000 concept detectors.
Similar to Chen \emph{et al.} \cite{shih-fu:2014} we rely on social tagged media, be it that we focus on tagged videos as also used by Habibian \emph{et al.}~\cite{HabibianICM2014} and Mazloom \emph{et al.} \cite{MazloomICMR15}. However, rather than building concept detectors from the tagged videos \cite{HabibianICM2014,shih-fu:2014,WuCVPR2014,MazloomICMR15}, we prefer to use the tags directly for video representation.

\subsection{Representations from tags}
We are inspired by recent progress in socially tagged image retrieval \cite{LiArxive15}, where many have demonstrated the value of tags for image retrieval. While it is well known that tags are often ambiguous, faulty, and incomplete, these limitations can be overcome to some extent by clever algorithms. Two representative and good performing~\cite{LiArxive15} algorithms are neighbor voting by Li \emph{et al.} \cite{xirong:tmm09} and TagProp by Guillaumin \emph{et al.} \cite{tagprop}. Given an image, the neighbor voting algorithm first retrieves its nearest neighbors from a source set in terms of low-level visual similarity. To determine the relevance of each tag of the input image, the algorithm then simply counts the tag's occurrence in annotations of the top-$k$ most similar images. Apart from tag refinement, the algorithm can also be leveraged for tag assignment. In this scenario the tags from the neighbors are sorted in descending order in terms of their occurrence frequency, and the top ranked tags are propagated to the input image. Different from the neighbor voting algorithm which considers the neighbors equally important, TagProp assigns rank-based or distance-based weights to the individual neighbors such that tags from neighbors closer to the input image will be enhanced in the tag propagation process. Our solution is grounded on tag propagation similar to \cite{xirong:tmm09,tagprop}, but takes two steps further to make it more suited for video event detection. One, instead of frame-level tag propagation as a straightforward application of \cite{xirong:tmm09,tagprop} to the video domain, we conduct video-level tag propagation. Since the number of videos is much smaller than the number of video frames, this design ensures good scalability of our solution to deal with large-scale video sets. Two, we conduct tag refinement on the weakly labeled training video set before using it as a resource for tag propagation. This resolves to some extent the inaccuracy and the incompleteness of social tags assigned to the source videos. As a consequence, more relevant tags will be propagated to the input video.

Propagating tags between videos has been studied in the context of tag recommendation \cite{Ballan:mm11,stefan:sigir09}. There, tags are meant to be used by end users, mainly for video browsing and retrieval. In contrast, we propagate tags for the purpose of using them as video representation for computing (cross-media) relevance between an unlabeled video and a specific event.

\subsection{Contributions}
Our work makes the following contributions. First of all, we propose a new semantic video representation for event detection using social tags that can be associated to videos. 
To the best of our knowledge, no method currently exists in the literature able to represent a video for event detection using just its tags, other than our previous conference paper \cite{Mazloom:ICMR14}. 
It should be noted that \cite{Mazloom:ICMR14} proposes a language model on top of the representation for video retrieval using query by zero, one or multiple positive examples. 
Here we prefer the parameter-free cosine distance for zero-shot event detection and exploit a support vector machine for the scenario where a few positive \textit{and} many negative video examples are available to learn an event classifier.
In addition, we introduce source set refinement, which differentiates between the tags of neighbor videos in advance to tag propagation.
Consequently, we obtain an improved bag of tags per video by considering source set refinement and multiple tag assignment functions. 
We show the merit of our proposal by performing several experiments on more than 1,000 hours of arbitrary Internet videos from the TRECVID Multimedia Event Detection task 2013, 2014 and the Columbia Consumer Video dataset. We call our approach \emph{TagBook}, and detail its construction for few-example and zero-example event detection next.

\section{TagBook based Video Event Detection} \label{sec:method}

\subsection{Problem Formalization} \label{ssec:problem}

Given a user specified event, video event detection is to retrieve videos showing the event from a large set of unlabeled videos. For the ease of consistent description, we use $e$ to indicate the given event, $v$ be a video, and $\mathcal{V}=\{v_1,\ldots,v_n\}$ a test set of $n$ videos. We aim to construct a real-valued function $f(v,e)$ which produces the relevance score between the video and the event. By sorting $\mathcal{V}$ according to $f(v,e)$ in descending order, videos most relevant with respect to the event will be obtained.

Let $\mathcal{V}_l=\{(v_{l,1},y_1),\ldots,(v_{l,p},y_p)\}$ be a set of $p$ labeled video samples available for a specific event, where $y_i=1$ means positive samples and $y_i=-1$ for negative samples. The difficulty in constructing $f(v,e)$ largely depends on the size of $\mathcal{V}_l$. Here the amount of positive samples is our concern, as the occurrence of a specific event in a video collection tends to be rare, making the acquisition of positive samples much more expensive than obtaining negative samples. In practice, even finding a single sample could be tricky, and one has no other choice than to express the event in words.

We now describe more formally the two scenarios of video event detection, in an order of increasing difficulty:
\begin{enumerate}
\item \textit{Few-example} video event detection: Finding videos relevant to a specific event $e$ from $\mathcal{V}$, given $|\mathcal{V}_l|>=1$.
Typically $\mathcal{V}_l$ has a handful of positive examples. 
\item \textit{Zero-example} video event detection: Finding videos relevant to a specific event $e$ from $\mathcal{V}$, given $\mathcal{V}_l=\emptyset$.
      In this case, the event is described by a natural language sentence $q$.
\end{enumerate}

The scarcity of video samples combined with the high dimensionality of low-level visual features makes it nontrivial to construct $f(v,e)$ effectively. Moreover, in the zero-example scenario, the visual features are inapplicable to compute cross-media similarity between a video and a description. To resolve these difficulties, we present TagBook, a compact and semantic representation of an entire video, which works for both scenarios.

The key idea of TagBook is to represent an unlabeled video $v$ by a fixed-length tag vector, denoted as $\mathbf{b}(v)$. Let $\mathcal{T}=\{t_1,\ldots,t_m\}$ be a vocabulary of $m$ distinct tags used in the TagBook. Each dimension of the tag vector uniquely corresponds to a specific tag, where $\mathbf{b}(v,i)$ is the relevance score between the tag $t_i$ and the video $v$. Hence, TagBook essentially embeds a video into an $m$-dimensional tag space.

Next, we show in Section \ref{ssec:use-tagbook} how to tackle video event detection using TagBook, followed by a solution to implement this representation in Section \ref{ssec:create-tagbook}. For the ease of reference, Table \ref{tab:notation} lists the main notation used throughout this work.

\begin{table}[tb!]
\centering \caption{Main notations defined in this work}
\label{tab:notation} \centering
 \scalebox{1}{
 \begin{tabular}{@{}ll@{}}
 \toprule
 \textbf{Notation} & \textbf{Definition} \\
 \midrule
  $v$ & a video\\
  $e$ & a video event \\
  $t$ & a tag \\
  $\mathbf{b}(v)$ & a tag vector of a given video \\
  $\mathbf{b}(e)$ & a tag vector of a given event \\
  $\mathbf{b}_{few}(e)$ & few-example version of $\mathbf{b}(e)$ \\
  $\mathbf{b}_{zero}(e)$  & zero-example version of $\mathbf{b}(e)$ \\
  $f(v,e)$ & a relevance function computed as $cosine(\mathbf{b}(v),\mathbf{b}(e))$\\
  $\mathcal{T}$ & a vocabulary of $m$ tags \\
  $\mathcal{V}$ & a set of unlabeled test videos \\
  $\mathcal{V}_l$ & a set of labeled video samples of a given event\\
  $\mathcal{V}_s$ & a set of socially tagged videos for tag propagation \\
  $s(v,v')$ & visual similarity between two videos\\
  $\llbracket v_s,t \rrbracket$ & a binary function indicating if $v_s \in \mathcal{V}_s$ is labeled with $t$ \\
  $r(v_s,t)$ & the relevance score between $v_s \in \mathcal{V}_s$ and tag $t$ \\

 \bottomrule
 \end{tabular}
 }
 \end{table}

\begin{figure}[!t]
\centering
\includegraphics[width=\columnwidth]{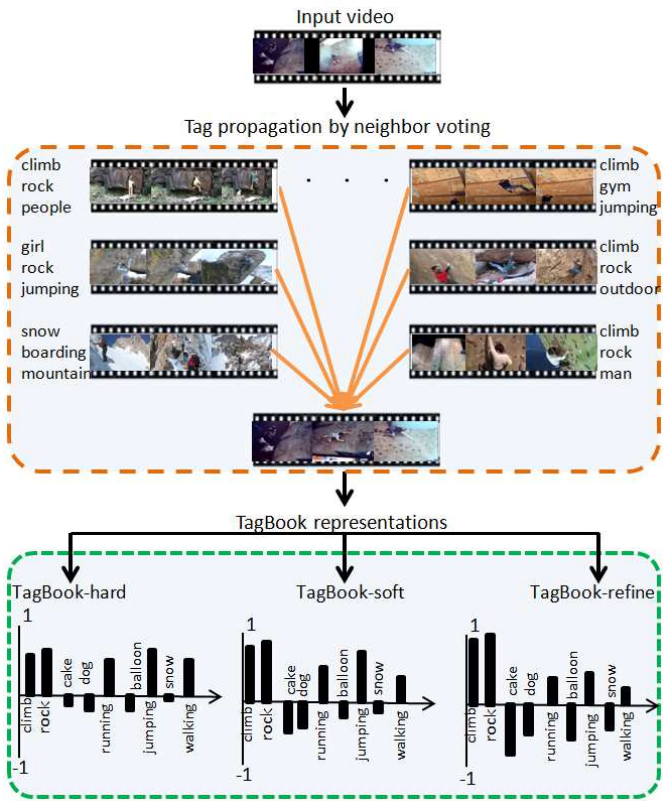}
\caption{A conceptual framework for generating a TagBook representation for an unlabeled video. 
From a source set of web videos annotated by online users we propagate tags from the visually similar video neighbors of the input video. 
Depending on how the neighbors are weighted and whether the tags of the source set are refined,
we derive three variants of TagBook,
i.e., TagBook-hard (equal neighbors and raw tags),
TagBook-soft (weighted neighbors and raw tags), and TagBook-refine (weighted neighbors and refined tags).
} \label{fig:method}
\end{figure}

\subsection{Two-Scenarios for Video Event Detection using TagBook} \label{ssec:use-tagbook}
We explain how a specific event $e$ can be represented as a TagBook. Let $\mathbf{b}(e)$ be the tag vector of an event. The relevance between this event and a video boils down to computing the cosine similarity between the two tag vectors. That is,
\begin{equation} \label{eq:relevance}
f(v,e) := cosine(\mathbf{b}(v), \mathbf{b}(e)).
\end{equation}
Notice that we have also investigated other similarity metrics including the Euclidean distance, the Spearman rank correlation, the Jensen--Shannon divergence, the $\chi^2$ distance, histogram intersection, and the Earth Mover's Distance. Among them, the cosine similarity strikes the best balance between effectiveness and efficiency. We use $\mathbf{b}_{few}(e)$ and $\mathbf{b}_{zero}(e)$ to indicate two variants corresponding to the few-example and zero-example scenarios, respectively.

In the few-example scenario, the event $e$ is expressed in terms of $p$ labeled video samples. Some of these samples could be more important than others for modeling the event. Hence, we consider the tag vector of the event as a weighted combination of its samples. In particular, we define:
\begin{equation} \label{eq:few-tagbook}
\mathbf{b}_{few}(e) := \sum_{i=1}^p \alpha_i y_i \mathbf{b}(v_{l,i}),
\end{equation}
where $\{\alpha_i\}$ are weight parameters. Notice that Eq. (\ref{eq:few-tagbook}) bears high resemble to the decision function of a linear Support Vector Machine (SVM). Hence, we optimize the weights by a linear SVM solver \cite{pegasos}.


In the zero-example scenario, a textual description $q$ of the event is provided. Using the classical bag-of-words model, $q$ is converted to a tag vector. Accordingly, $\mathbf{b}_{zero}(e,i)$ is 1 if $t_i$ is in $q$, and $0$ otherwise.

With $\mathbf{b}(e)$ in  Eq. (\ref{eq:relevance}) replaced by $\mathbf{b}_{few}(e)$ and $\mathbf{b}_{zero}(e)$ separately, we have the relevance functions $f_{few}(x,e)$ and $f_{zero}(x,e)$ for each of the two scenarios.

\subsection{TagBook Construction by Content-Based Tag Propagation} \label{ssec:create-tagbook}
We propose to construct the TagBook representation of an unlabeled video by propagating tags from a large set of $N$ socially tagged videos, denoted by $\mathcal{V}_s = \{v_{s,1}, \ldots, v_{s,N}\}$. Each video $v_s \in \mathcal{V}_s$ is assigned with a limited number of social tags. For each tag $t \in \mathcal{T}$, we use a binary labeling function $\llbracket v_s,t \rrbracket$, which outputs $1$ if $v_s$ is labeled with $t$, and $0$ otherwise. Due to the subjective nature of social tagging, some of the assigned tags could be irrelevant with respect to the visual content of $v_s$.

With the hypothesis that visually similar images shall have similar tags, content-based tag propagation has been exploited in the context of image auto-tagging \cite{xirong:tmm09,tagprop}. Tags are propagated from neighbor images which are visually close to a test image, where the neighbors are treated either equally \cite{xirong:tmm09} or weighted in terms of their visual distance to the test image \cite{tagprop}. In our context, let $\{\hat{v}_{s,1}, \ldots, \hat{v}_{s,k}\}$ be the $k$ nearest neighbor videos retrieved from $\mathcal{V}_s$ by a predefined video similarity $s(v,v')$. A general formula of tag propagation can be expressed as 
\begin{equation} \label{eq:tagprop}
\mathbf{b}(v, i) = \frac{1}{k}\sum_{j=1}^k s(v, \hat{v}_{s,j}) \cdot r(\hat{v}_{s,j},t_i),
\end{equation}
where $r(v_s,t)$ measures the relevance of a specific tag $t$ with respect to a specific video $v_s \in \mathcal{V}_s$.
To simplify our notation, we abuse $s(v,v')$ to let it also indicate the contribution of a neighbor video in the tag propagation process. For instance, in a hard assignment mode, the output of $s(v,v')$ will be binary, producing 1 if the rank of the neighbor is within k, and 0 otherwise.
%
Tags of higher occurrence in $\mathcal{V}_s$ are more likely to be propagated. In order to reduce such an effect, we subtract $\mathbf{b}(v, i)$ by
a term related to tag occurrence, i.e.,
\begin{eqnarray} \label{eq:tagprop-prior}
\mathbf{b}(v, i)  && =    \frac{1}{k}\sum_{j=1}^k s(v, \hat{v}_{s,j}) \cdot r(\hat{v}_{s,j},t_i) \nonumber \\
         && - \frac{1}{N}\sum_{j=1}^N s(v, v_{s,j}) \cdot r(v_{s,j},t_i)
\end{eqnarray}

Concerning $r(v_s,t)$, a straightforward choice is to instantiate it using the labeling function $\llbracket v_s,t \rrbracket$. As aforementioned, this choice is questionable due to the inaccuracy and sparseness of social tags. We therefore conduct tag refinement on the source set $\mathcal{V}_s$ \emph{before} using it for TagBook construction. Again, tag propagation is employed, computing $r(v_s,t)$ by
\begin{eqnarray} \label{eq:tag-refine}
r(v_s,t) & = & \frac{1}{k_r} \sum_{j=1}^{k_r} s(v_s, \tilde{v}_{s,j}) \cdot \llbracket \tilde{v}_{s,j},t \rrbracket \nonumber \\
         && - \frac{1}{N} \sum_{j=1}^N s(v_s, v_{s,j}) \cdot \llbracket v_{s,j},t \rrbracket
\end{eqnarray}
where $\{\tilde{v}_{s,1}, \ldots, \tilde{v}_{s,k}\}$ are the $k_r$ nearest neighbors of $v_s$ retrieved from the source set. Both $k$ and $k_r$ are empirically set to be 500.

Concerning the content-based similarity between two videos $s(v,v')$, we use convolutional neural network (CNN) features for their well recognized performance. In particular, we train the AlexNet \cite{KrizhevskySH12} for over 15k ImageNet classes, each having at least 50 positive examples. Given a video, we extract its frames uniformly with a time interval of two seconds. The second fully connected layer (FC2) is used, representing each frame with a 4,096-dimensional feature vector. The video-level feature vector is obtained by average pooling over all the frame-level vectors.
The video similarity $s(v,v')$ is computed as the cosine similarity between the corresponding CNN feature vectors.

Depending on how the weights of the neighbors and $r(v_s,t)$ are implemented, we present three variants of TagBook's tag assignment, namely
\begin{enumerate}
\item TagBook-hard: Neighbor videos are assigned with binary weights, i.e., 1 if the rank of the neighbor is within k, and 0 otherwise, and $r(v_s,t)$ as $\llbracket v_s,t \rrbracket$.
\item TagBook-soft: Neighbor videos are weighted in terms of their similarity scores, and $r(v_s,t)$ as $\llbracket v_s,t \rrbracket$. TagBook-soft corresponds to the representation used in \cite{Mazloom:ICMR14}.
\item TagBook-refine: Neighbor videos are weighted in terms of their similarity scores, and $r(v_s,t)$ as Eq. (\ref{eq:tag-refine}).
\end{enumerate}

Fig.~\ref{fig:method} illustrates the TagBook generation process. Next, we evaluate TagBook-hard, TagBook-soft, and TagBook-refine for video event detection on three benchmark datasets.

\section{Evaluation} \label{sec:eval}

\subsection{Datasets} \label{ssec:datasets}
\textbf{Source set}. As our social-tagged video collection $\mathcal{V}_s$, we adopt the VideoStory46K dataset from Habibian \emph{et al.}~\cite{HabibianICM2014} which contains 46k videos from YouTube. Every video has a short caption provided by the person who uploaded the video. From the captions we remove stop words and words not visually detectable such as God (we used the visualness filter from~\cite{HabibianICM2014}) and finally obtain a vocabulary $\mathcal{T}$ of 19,159 unique tags.

\textbf{Test datasets 1 and 2: TRECVID MED 2013 and MED 2014}~\cite{MED13}. The MED corpus contains user-generated web videos with a large variation in quality, length and content of real-world events including life events, instructional events, sport events, etc. Both the 2013 and 2014 corpus consist of several partitions with ground truth annotation at video-level for 30 event categories, with 10 of those events overlapping in both 2013 and 2014. 
For the few-example scenario, we follow the TRECVID 10Ex evaluation procedure~\cite{2013trecvidover}. That is, for each event its training data $\mathcal{V}_l$ contains 10 positive video samples from the Event Kit training data, and 5K negative video samples from the Background training data. In the zero-example scenario,  we rely only on the TRECVID provided textual definition of a test event. For both scenarios we report results on the MED 2013 test set and the MED 2014 test set, each containing 27K videos.

\textbf{Test dataset 3: Columbia CV}~\cite{icmr11:CCV}. This corpus consists of 9,317 YouTube videos, and crowd-sourced ground truth with respect to 20 visual concepts. Fifteen of the concepts correspond to specific events such as \emph{Ice skating}, \emph{Birthday}, and \emph{Music performance}, so only these event-related concepts are considered in our experiments. We start from the official data partition, i.e., a training set of 4,625 videos and a test set of 4,637 videos. 
For few-example event detection, similar to Habibian et al. \cite{HabibianICM2014} we down-sample the training set to have at most 10 positive training examples per event, obtained based on the alphabetical order of the video names. Different from the TRECVID datasets, the Columbia CV dataset does not provide textual definition of events. So we do not perform zero-example video event detection on this dataset.

In what follows, we first use the MED 2013 dataset to find a good implementation of TagBook, achieved by evaluating varied choices including refining annotations of the source set, visual neighbor re-weighting, and the TagBook size. 
To study whether a more complex model with higher non-linear capability would help improve the accuracy of event detection,
we compare in the few-example setting the linear model and a non-linear variant, reporting both speed and accuracy.
To assess how the learned implementation generalizes to new test data, we evaluate it using the other two test sets, with a comparison to several state-of-the-art video representations.

As performance metrics, Average Precision (AP) per event and Mean Average Precision (MAP) per dataset are reported.

\begin{table*}[t!] \centering \caption{Comparing three variants of TagBook on TRECVID MED 2013. Full-size TagBooks are used. For each scenario, top performers per event are highlighted in bold font.}
\label{tab:res1} \centering
\scalebox{1}{
\begin{tabular}{@{}lccc c ccc@{}}
\toprule
 &  \multicolumn{3}{c}{\textbf{Zero-example}} && \multicolumn{3}{c}{\textbf{Few-example}}  \\
 \cmidrule{2-4}  \cmidrule{6-8}
 \textbf{Event}                      & TagBook-hard & TagBook-soft & TagBook-refine && TagBook-hard & TagBook-soft & TagBook-refine \\
 \midrule
  Birthday party                     & 0.028          & 0.051          & \textbf{0.065}       && 0.099          & 0.136 & \textbf{0.149} \\
  Changing a vehicle tire            & 0.081          & 0.108          & \textbf{0.125}       && 0.278          & 0.402 & \textbf{0.466} \\
  Flash mob gathering                & 0.145          & 0.194          & \textbf{0.221}       && 0.294          & 0.372 & \textbf{0.399} \\
  Getting a vehicle unstuck          & 0.198          & 0.211          & \textbf{0.235}       && 0.499          & 0.547 & \textbf{0.587} \\
  Grooming an animal                 & 0.046          & 0.066          & \textbf{0.095}       && 0.101          & 0.165 & \textbf{0.201} \\
  Making a sandwich                  & 0.019          & 0.021          & \textbf{0.036}       && 0.038          & 0.040 & \textbf{0.076} \\
  Parade                             & 0.192          & 0.201          & \textbf{0.204}       && 0.210          & 0.228 & \textbf{0.230} \\
  Parkour                            & 0.094          & 0.100          & \textbf{0.109}       && 0.229          & 0.308 & \textbf{0.334} \\
  Repairing an appliance             & 0.200          & 0.277          & \textbf{0.298}       && 0.256          & 0.376 & \textbf{0.381} \\
  Working on a sewing project        & \textbf{0.034} & 0.031          &  0.027               && \textbf{0.087} & 0.066 & 0.072 \\
  Attempting a bike trick            & 0.029          & 0.067          & \textbf{0.087}       && 0.083          & 0.146 & \textbf{0.199} \\
  Cleaning an appliance              & 0.008          & 0.004          & \textbf{0.019}       && 0.016          & 0.009 & \textbf{0.028} \\
  Dog show                           & \textbf{0.125} & 0.084          & 0.091                && \textbf{0.232} & 0.121 & 0.143 \\
  Giving directions to a location    & 0.003          & \textbf{0.006} & \textbf{0.006}       && 0.006          & 0.008 & \textbf{0.009} \\
  Marriage proposal                  & 0.002          & 0.002          & \textbf{0.003}       && 0.005          & 0.007 & \textbf{0.009} \\
  Renovating a home                  & 0.008          & 0.010          & \textbf{0.013}       && 0.023          & 0.028 & \textbf{0.044} \\
  Rock climbing                      & \textbf{0.043} & 0.021          & 0.026                && \textbf{0.124} & 0.084 & 0.098 \\
  Town hall meeting                  & \textbf{0.094} & 0.071          & 0.077                && \textbf{0.182} & 0.165 & 0.177 \\
  Winning a race without a vehicle   & 0.011          & 0.055          & \textbf{0.071}       && 0.171          & 0.232 & \textbf{0.275} \\
  Working on a metal crafts project  & 0.003          & 0.006          & \textbf{0.007}       && 0.014          & 0.050 & \textbf{0.067} \\
  \midrule
  \textbf{MAP}                       & 0.068          & 0.079          & \textbf{0.091}       && 0.148          & 0.174 & \textbf{0.198} \\
 \bottomrule
\end{tabular}
}
\end{table*}

\subsection{Experiment 1: Finding a Good TagBook}
Table \ref{tab:res1} gives the performance of TagBook-hard, TagBook-soft, and TagBook-refine on the MED 2013 test set. TagBook-soft performs better than TagBook-hard, with 0.079 \emph{versus} 0.068 for zero-example video event detection and 0.174 \emph{versus} 0.148 in the few-example scenario. TagBook-refine performs the best, scoring MAP of 0.091 and 0.198 in zero-example and few-example, respectively. Recall that the only difference between TagBook-hard and TagBook-soft is that the latter re-weights neighbor videos in terms of their visual similarity to a test video, and the only difference between TagBook-soft and TagBook-refine is that the latter uses enriched annotations of the source set. The result shows the joint use of source set refinement and neighbor re-weighting is beneficial for extracting a better TagBook representation from unlabeled videos.

\begin{figure}[!tb]
\centering
\includegraphics[width=\columnwidth]{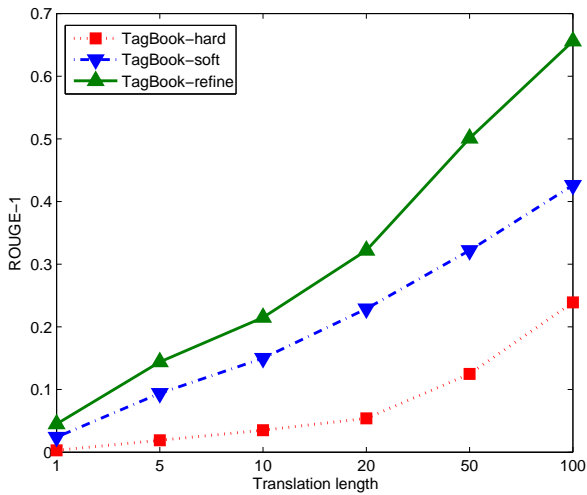}
\caption{TagBook-refine vs Other TagBooks for video description generation, tested on TRECVID MED 2013. TagBook-refined generates more accurate descriptions.} \label{fig:translation}
\end{figure}

\begin{figure}[!tb]
\centering
\includegraphics[width=\columnwidth]{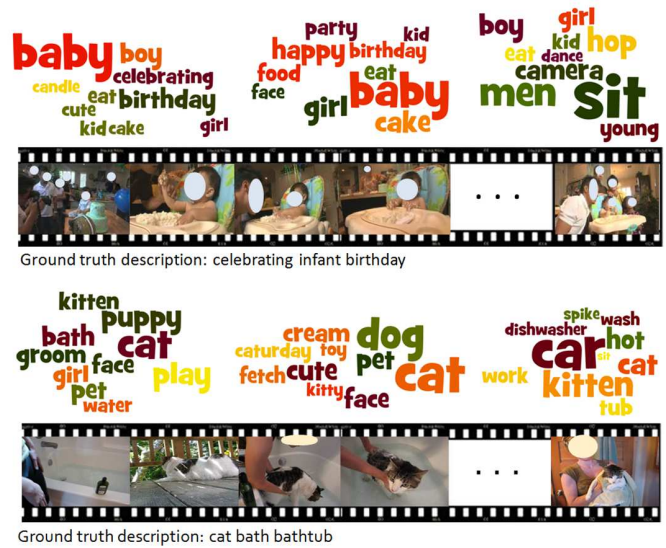}
\caption{Video examples along with their expert-provided description for the events \emph{birthday party} (top) and \emph{grooming an animal}. Tags predicted by TagBook-refine (left), TagBook-soft (middle), and TagBook-hard (right) are summarized as tag clouds. Tags generated by TagBook-refine tend to result in the best overlap with the ground truth (see Fig.~\ref{fig:translation}).} \label{fig:example-translation}
\end{figure}

We make a further comparison between the three variants of TagBook to see how well they describe a video. Given a test video, its TagBook based description is automatically generated by sorting tags in terms of their $\mathbf{b}(v,i)$ and keeping the top $\kappa$ ranked tags. We report the result of video description generation on the positive videos of each event class for which expert-provided descriptions are available. Following the protocol of~\cite{HabibianICM2014}, we use ROUGE-1, a performance metric computing the recall of the ground truth words in the generated description, thus increasing along with $\kappa$. The performance curve is shown in Fig.~\ref{fig:translation}, with real examples in Fig.~\ref{fig:example-translation}. Both figures demonstrate that TagBook-refine generates more accurate video descriptions.

\begin{figure}[!tb]
\centering
 \subfigure[Few-example video event detection] {
\noindent\includegraphics[width=\columnwidth]{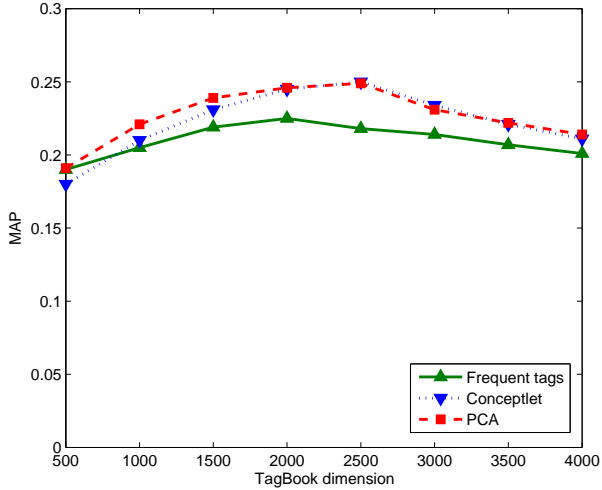}
\label{fig:tagbook-size-few}}
 \subfigure[Zero-example video event detection] {
\noindent\includegraphics[width=\columnwidth]{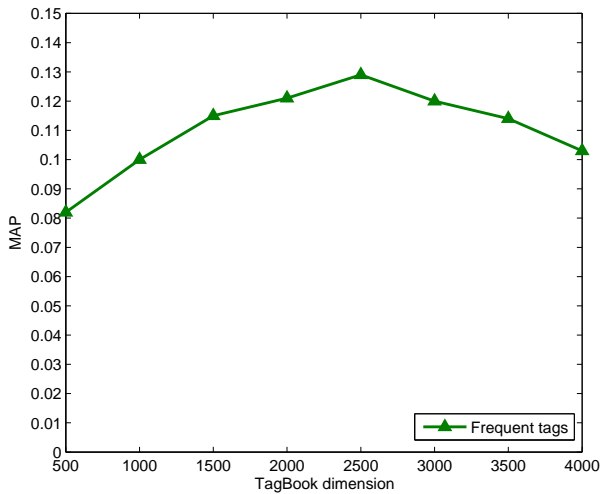}
\label{fig:tagbook-size-zero}}
\caption{The influence of the TagBook size on (a) few-example and (b) zero-example video event detection. Compared to the full-sized TagBook, TagBooks consisting of around 2,000 most frequent tags yield the best performance on the TRECVID MED 2013 test set. Since Conceptlet and PCA require visual examples, they are inapplicable in the zero-example scenario.}
\label{fig:tagbook-size}
\end{figure}

To assess the effect of the TagBook size on video event detection performance, we investigate three dimension reduction methods. The first and the most straightforward method is to preserve the top frequent tags in the source set. We term it \emph{Frequent tags}. The second is the classical Principal Component Analysis (PCA). The last is Conceptlet~\cite{Mazloom:TMM2014}, a state-of-the-art concept selection algorithm, aiming for the best subset of concepts per event by considering correlations between concepts. Notice that PCA and Conceptlet require positive video examples, making them inapplicable in the zero-example scenario. As shown in Figure \ref{fig:tagbook-size}, for all the three methods, size-reduced TagBooks score higher MAP than the full-sized TagBook. In particular, peak performance is reached at the size of 2,000 for the few-example case, and 2,500 for the one-example case. In the remaining part of the evaluation, we use TagBook-refine reduced by the \emph{Frequent tags} method, for its good performance, simplicity, and applicability for both scenarios.

Finally, we also assessed the scenario where TagBook-refine relies on a non-linear $\chi^2$ feature embedding \cite{pami2012-efm} rather than a linear kernel. 
As Table \ref{tab:nonlinear} shows, few-example event detection with TagBook-refine profits from a non-linear kernel at the expense of an increased computation time. 
The mean average precision tends to be about 10\% higher for the non-linear kernel, but computation is also about ten times as much.
In the remaining few-example event detection experiments we rely on the linear kernel for its good accuracy and efficiency tradeoff.

\begin{table}[tb!]
\centering \caption{Linear vs non-linear kernel, using TagBook-refine with varying sizes, for few-example event detection on TRECVID MED 2013. 
We report the time needed for training models and testing them for all events,
measured in seconds on Intel Xeon Processor E5-2690.
The non-linear kernel is more effective at the expense of an almost ten-fold increase in computation time on average.
} \label{tab:nonlinear} \centering
 \scalebox{0.8}{
 \begin{tabular}{@{}l rrr c rrr@{}}
\toprule 
          & \multicolumn{3}{c}{\textbf{Linear}} && \multicolumn{3}{c}{\textbf{Nonlinear}} \\
          \cmidrule{2-4} \cmidrule{6-8}
 TagBook size     & MAP & Training time & Test time && MAP & Training time & Test time \\
 \cmidrule{1-8} 
1,000 & 0.209 &  9.01 &  6.00 && 0.224 &  55.11  & 39.89 \\
1,500 & 0.219 &  9.88 &  7.12 && 0.235 &  81.00  & 69.00 \\
2,000 & 0.225 & 14.02 &  8.98 && 0.244 & 114.03 &  85.98 \\
2,500 & 0.218 & 16.01 & 10.99 && 0.247 & 141.86 & 116.14 \\
 \bottomrule
 \end{tabular}
}
\end{table}
%

\subsection{Experiment 2: TagBook versus Others}
We compare TagBook with several state-of-the-art video representations for event detection.

\emph{1. CNN-FC2}. This representation has been described in Section \ref{sec:method} for finding similar videos. 

\emph{2. ConceptVec-15k}. For each sampled frame of a specific video, instead of the CNN-FC2 layer we adopt the output of the AlexNet's softmax layer. The output is a nonnegative vector, where each dimension corresponds to one of the 15k ImageNet concepts and its value is a probabilistic estimation of the concept present in the frame. Average pooling is used to obtain the video-level representation.

\emph{3. ConceptVec-2k.} As aforementioned, the TagBook is essentially constructed by neighbor voting based on tag propagation. One might consider using more advanced mode-based techniques such as SVMs. To address this concern, for each of the top 2k most frequent tags in our source set, we learn a separate linear SVMs classifier with CNN-FC2 as the underlying feature. A video in the source set is taken as positive training examples if its caption contains the tag, and used as negatives otherwise. By applying the classifiers, each video is represented by a 2k vector of concept detector outputs.

\emph{4. VideoStory}~\cite{HabibianICM2014}. This video event representation strives to embed the caption of a video and its visual features in a joint space by grouping tags. We follow the author suggested implementation~\cite{HabibianICM2014}, which encodes each video as a Fisher vector over MBH descriptors along the motion trajectories. We learn the joint embedding from the source set with an optimal target dimensionality of 2,048.

\emph{5. Concept Prototypes}~\cite{MazloomICMR15}. Video event representation that learns a set of relevant frames as the concept prototypes and uses the prototypes for representing a video. We follow the author suggested implementation~\cite{MazloomICMR15}, which first encodes each video frame as CNN-FC2, and maps it to a concept prototype space learned for 479 concepts.

Except for CNN-FC2, all video representations are semantic and can therefore be used in both few-example and zero-example scenarios. For fair comparisons, the same event modeling technique, i.e., Eq. (\ref{eq:few-tagbook}), is used, making the choice of video representation the only variable. This setting allows us to precisely identify which representation is the best.



The performance of video event detectors built on the varied representations is summarized in Table \ref{tab:med13}, Table \ref{tab:med14}, and Table \ref{tab:ccv}, corresponding to TRECVID MED 2013, TRECVID MED 2014, and CCV, respectively. We directly cite AP scores from the original papers whenever applicable. Consequently, the results of VideoStory and Concept Prototypes are only partially available.

\begin{table*}[tb!]
 \label{tab:med13} \centering \caption{TagBook versus others on  TRECVID MED 2013.} 
\scalebox{0.66}{
\begin{tabular}{@{}lcccccc c cccc@{}}
\toprule
                     & \multicolumn{6}{c}{\textbf{Few-example}} && \multicolumn{4}{c}{\textbf{Zero-example}} \\
                     \cmidrule{2-7} \cmidrule{9-12}
\textbf{Event}        & \textbf{CNN-FC2} & \textbf{ConceptVec-15k} & \textbf{ConceptVec-2k} & \textbf{VideoStory}~\cite{HabibianICM2014} & \textbf{Concept Prototypes}~\cite{MazloomICMR15}   &  \textbf{TagBook} && \textbf{ConceptVec-15k} & \textbf{ConceptVec-2k} &  \textbf{Concept Prototypes}~\cite{MazloomICMR15}  & \textbf{TagBook} \\
\midrule
Birthday party                     & 0.137  & 0.114  & 0.156  & 0.118 & \textbf{0.188} & 0.182  && 0.022  & 0.075  & 0.154 & \textbf{0.155}\\ 
Changing a vehicle tire            & 0.391  & 0.388  & 0.411  & 0.103 & 0.464 & \textbf{0.560}  && 0.099  & 0.181  & 0.320 & \textbf{0.337}\\ 
Flash mob gathering                & 0.405  & 0.347  & 0.421  & \textbf{0.535} & 0.439 & 0.317  && 0.104  & 0.178  & \textbf{0.271} & 0.174\\  
Getting a vehicle unstuck          & 0.334  & 0.323  & 0.456  & 0.319 & 0.418 & \textbf{0.602}  && 0.107  & 0.201  & \textbf{0.406} & 0.312\\
Grooming an animal                 & 0.084  & 0.108  & 0.149  & 0.151 & 0.154 & \textbf{0.247}  && 0.019  & 0.101  & 0.095 & \textbf{0.201}\\
Making a sandwich                  & 0.031  & 0.074  & 0.087  & 0.074 & \textbf{0.131} & 0.108  && 0.021  & 0.031  & \textbf{0.164} & 0.099\\
Parade                             & 0.171  & 0.109  & 0.271  & \textbf{0.452} & 0.303 & 0.279  && 0.094  & 0.135  & \textbf{0.240} & 0.185\\
Parkour                            & 0.330  & 0.309  & 0.378  & \textbf{0.721} & 0.326 & 0.467  && 0.020  & 0.131  & 0.112 & \textbf{0.215}\\
Repairing an appliance             & 0.169  & 0.127  & 0.261  & 0.184 & 0.244 & \textbf{0.395}  && 0.078  & 0.157  & \textbf{0.213} & 0.211\\
Working on a sewing project        & 0.058  & 0.071  & 0.107  & \textbf{0.151} & 0.109 & 0.126  && 0.016  & 0.036  & 0.089 & \textbf{0.098}\\
Attempting a bike trick            & 0.054  & 0.030  & 0.123  & 0.061 & 0.144 & \textbf{0.200}  && 0.017  & \textbf{0.067}  & 0.061 & 0.066\\
Cleaning an appliance              & 0.021  & 0.019  & 0.035  & \textbf{0.078} & 0.055 & 0.038  && 0.006  & 0.019  & \textbf{0.026} & 0.023\\
Dog show                           & 0.232  & 0.134  & 0.254  & \textbf{0.354} & 0.313 & 0.243  && 0.003  & 0.155  & 0.011 & \textbf{0.200}\\
Giving directions to a location    & 0.012  & 0.005  & 0.011  & 0.004 & \textbf{0.022} & 0.013  && 0.004  & 0.004  & \textbf{0.008} & 0.005\\
Marriage proposal                  & 0.002  & 0.002  & \textbf{0.009}  & 0.004 & 0.004 & 0.007  && \textbf{0.004}  & 0.002  & 0.005 & 0.003\\
Renovating a home                  & 0.019  & 0.024  & 0.046  & \textbf{0.051} & 0.033 & 0.053  && 0.017  & 0.011  & \textbf{0.026} & 0.018\\
Rock climbing                      & 0.070  & 0.063  & \textbf{0.127}  & 0.100 & 0.110 & 0.097  && 0.003  & 0.020  & \textbf{0.036} & 0.026\\
Town hall meeting                  & 0.268  & 0.201  & 0.200  & 0.118 & \textbf{0.290} & 0.236  && 0.008  & 0.087  & 0.035 & \textbf{0.148}\\
Winning a race without a vehicle   & 0.150  & 0.126  & 0.153  & 0.217 & 0.182 & \textbf{0.245}  && 0.012  & 0.045  & \textbf{0.101} & 0.099\\
Working on a metal crafts project  & 0.054  & 0.068  & 0.099  & 0.118 & \textbf{0.144} & 0.079  && 0.002  & 0.005  & \textbf{0.014} & 0.002\\
\midrule
\textbf{MAP}                       & 0.150  & 0.132  & 0.188  & 0.196 & 0.204 & \textbf{0.225}  && 0.032  & 0.081  & 0.119 & \textbf{0.129}\\
\bottomrule
\end{tabular}
}
\end{table*}

\begin{table*}[tb!]
\centering \caption{TagBook versus others on TRECVID MED 2014.} \label{tab:med14} \centering
\scalebox{1}{
\begin{tabular}{@{}lcccc c ccc@{}}
\toprule
                     & \multicolumn{4}{c}{\textbf{Few-example}} && \multicolumn{3}{c}{\textbf{Zero-example}} \\
                     \cmidrule{2-5} \cmidrule{7-9}
\textbf{Event}       & CNN-FC2 & ConceptVec-15k & ConceptVec-2k &  TagBook && ConceptVec-15k & ConceptVec-2k &  TagBook \\
\midrule
Attempting a bike trick           &  0.057    &  0.127  &  0.134  & \textbf{0.139} && 0.016    &  0.042  &  \textbf{0.075}  \\
Cleaning an appliance             &  0.022    &  0.062  &  0.072  & \textbf{0.119} && 0.014    &  0.071  &  \textbf{0.080} \\
Dog show                          &  0.215    &  \textbf{0.361}  &  0.271  & 0.312 && 0.016    &  \textbf{0.162}  &  0.157   \\
Giving directions to a location   &  0.013    &  \textbf{0.051}  &  0.030  & 0.032 && 0.003    &  0.004  &  \textbf{0.006}  \\
Marriage proposal                 &  0.003    &  0.005  &  \textbf{0.008}  & \textbf{0.008} && \textbf{0.008}    &  0.005  &  0.005\\
Renovating a home                 &  0.022    &  0.050  &  0.050  & \textbf{0.083} &&  0.016    &  0.046  &  \textbf{0.047} \\
Rock climbing                     &  0.066    &  0.061  &  \textbf{0.101}  & 0.089 &&  0.005    &  0.008  &  \textbf{0.020} \\
Town hall meeting                 &  \textbf{0.268}    &  0.212  &  0.204  & 0.228 &&  0.008    &  0.102  &  \textbf{0.120} \\
Winning a race without a vehicle  &  0.126    &  0.121  &  0.130  & \textbf{0.175} &&  0.017    &  \textbf{0.087}  &  0.063 \\
Working on a metal crafts project &  0.038    &  0.037  &  \textbf{0.082}  & 0.072 &&  0.003    &  \textbf{0.006}  &  0.005 \\
Beekeeping                        &  0.410    &  \textbf{0.525}  &  0.461  & 0.502 &&  \textbf{0.062}    &  0.003  &  0.009 \\
Wedding shower                    &  0.074    &  0.044  &  0.072  & \textbf{0.117} &&  0.021    &  0.012  &  \textbf{0.035} \\
Non-motorized vehicle repair      &  0.228    &  0.407  &  \textbf{0.415}  & 0.398 && 0.003    &  0.074  &  \textbf{0.265} \\
Fixing musical instrument         &  0.077    &  0.085  &  0.097  & \textbf{0.114} && 0.008    &  0.003  &  \textbf{0.009} \\
Horse riding competetion          &  0.390    &  0.280  &  0.344  & \textbf{0.392} && \textbf{0.124}    &  0.075  &  0.118\\
Felling a tree                    &  0.030    &  0.100  &  0.086  & \textbf{0.118} && 0.006    &  0.021  &  \textbf{0.072} \\
Parking a vehicle                 &  0.111    &  \textbf{0.231}  &  0.088  & 0.119 && \textbf{0.198}    &  0.011  &  0.035 \\
Playing fetch                     &  0.007    &  0.033  &  0.031  & \textbf{0.055} && 0.005    &  0.020  &  \textbf{0.035} \\
Tailgating                        &  0.110    &  0.149  &  0.136  & \textbf{0.176} && 0.005    &  0.003  &  \textbf{0.006} \\
Tuning musical instrument         &  0.040    &  \textbf{0.079}  &  0.055  & \textbf{0.079} && \textbf{0.022}    &  0.006  &  0.009 \\
\midrule
\textbf{MAP}                      &  0.115    &  0.151  &  0.141  & \textbf{0.166} && 0.028    &  0.038  &  \textbf{0.059} \\
\bottomrule
\end{tabular}
}
\end{table*}

TagBook outperforms its competitors on all the three test datasets. In particular, as TagBook is built on top of CNN-FC2, its superior performance shows that TagBook is a more compact yet more semantic enriched video representation than the CNN feature.

For the model-based video representations, we observe that ConceptVec-2k is better than ConceptVec-15k in general. The main reason is that the ImageNet classes emphasize image objects, many of which are fine-grained classes of animals and plants. They are not meant for describing video events. By contrast, the source set from~\cite{HabibianICM2014} was collected from YouTube using event-like descriptions as queries. Learned from such data, ConceptVec-2k is more suited than ConceptVec-15k for video event detection.

TagBook is better than ConceptVec-2k, although they use the same source set and the same visual feature as their starting point. The main technical difference between TagBook and ConceptVec-2k is that the former is built in a model-free manner while the latter is model-based. User tags are known to be subjective and ambiguous, meaning large divergence in their imagery. Model-free approaches as neighbor voting can figure out a decision boundary much more complex than linear classifiers, making it more suited for addressing subjective tags. Besides, model-based approaches are more sensitive to noise. Because of these reasons, model-free approaches like neighbor voting are more effective for learning from user-tagged video data.

TagBook also compares favorably against VideoStory~\cite{HabibianICM2014} and Concept Prototypes~\cite{MazloomICMR15}. Recall that they all use the VideoStory46K dataset as their source set. Both VideoStory and Concept Prototypes trust the (weak) annotations and learn their representation directly on top of the dataset. TagBook, in contrast, enriches the source set first by suppressing noise and generating more relevant tags per video. Moreover, TagBook considers a weight per tag, rather than a binary presence or absence value.

We also compared against our previous work \cite{Mazloom:ICMR14}, which relies on TagBook-soft and a language model for retrieval. For fair comparison we use the same CNN features and the same source set. 
On TRECVID MED 2013, TagBook-refine improves over  \cite{Mazloom:ICMR14} for few-example event detection from 0.221 to 0.225 and for zero-example from 0.113 to 0.129

Finally, we make a system level comparison between the proposed TagBook based system and several state-of-the-art alternatives for zero-example video event detection. The results shown in Table~\ref{tab:zero-sota} again confirms the effectiveness of the TagBook as a new video representation for event detection.
In the most recent TRECVID MED evaluations from 2014 and 2015, other approaches have proven effective for few-example and zero-example event detection as well \cite{med2014:cmu,med2015:uva}. In \cite{med2014:cmu} the Informedia team from Carnegie Mellon University showed how a mixture of multimodal features, concepts and fusion schemes leads to state-of-the-art few-example event detection results. In addition, they have repeatedly demonstrated that pseudo-relevance feedback improves zero-example event detection \cite{mm2015jiang-cmu}. In \cite{med2015:uva} the MediaMill team from the University of Amsterdam proposed a better CNN video feature by enriched pretraining, leading to state-of-the-art results for few-example event detection. 
The PROGRESS set used in the TRECVID MED benchmark is for blind testing by NIST only, so we cannot compare directly, 
but we note that TagBook will profit from more discriminative and multimodal representations as well and is orthogonal to pseudo-relevance feedback.

\begin{table}[tb!]
\centering \caption{TagBook versus others on Columbia Consumer Video.} \label{tab:ccv} \centering
 \scalebox{0.73}{
 \begin{tabular}{@{}lccccc@{}}
\toprule 
\textbf{Event}           & \textbf{CNN-FC2} & \textbf{ConceptVec-15k} & \textbf{ConceptVec-2k} & \textbf{VideoStory} & \textbf{TagBook} \\
\midrule
 Basketball              & 0.466 & 0.515 & 0.547 & 0.553 & \textbf{0.633}  \\
 Baseball                & 0.551 & \textbf{0.608} & 0.563 & 0.299 & 0.594  \\
 Soccer                  & 0.507 & 0.504 & 0.546 & 0.505 & \textbf{0.574}  \\
 Ice skating             & 0.580 & 0.700 & \textbf{0.769} & 0.675 & 0.722  \\
 Skiing                  & 0.745 & 0.794 & \textbf{0.796} & 0.671 & \textbf{0.796}  \\
 Swimming                & 0.719 & 0.665 & 0.755 & \textbf{0.764} & 0.762  \\
 Biking                  & 0.435 & 0.435 & 0.507 & 0.561 & \textbf{0.621}  \\
 Graduation              & 0.261 & \textbf{0.295} & 0.278 & 0.121 & 0.290  \\
 Birthday                & 0.330 & 0.292 & \textbf{0.502} & 0.257 & 0.492  \\
 Wedding reception       & \textbf{0.214} & 0.174 & 0.161 & 0.117 & 0.196  \\
 Wedding ceremony        & \textbf{0.463} & 0.412 & 0.439 & 0.324 & 0.454  \\
 Wedding dance           & 0.399 & 0.296 & 0.423 & \textbf{0.521} & 0.503  \\
 Music performance       & 0.291 & 0.317 & 0.289 & 0.201 & \textbf{0.385}  \\
 Non-music performance   & 0.188 & 0.240 & 0.226 & 0.282 & \textbf{0.289}  \\
 Parade                  & 0.487 & 0.354 & 0.512 & \textbf{0.634} & 0.521  \\
 \midrule
 \textbf{MAP}            & 0.442 & 0.440 & 0.487 & 0.432 & \textbf{0.522}  \\
 \bottomrule
 \end{tabular}
}
\end{table}

\begin{table}[tb!]
\centering \caption{A system level comparison to the state-of-art for zero-example video event detection on TRECVID MED 2013.} \label{tab:zero-sota} \centering
 \scalebox{1}{
 \begin{tabular}{@{}lr@{}}
\toprule
\textbf{System}    &  MAP \\
\midrule
 Chen \emph{et al.}~\cite{shih-fu:2014}          & 0.024 \\
 Habibian \emph{et al.}~\cite{habibian14icmr}    & 0.063 \\
 Ye \emph{et al.}~\cite{ye2015}                  & 0.089 \\
 Chang \emph{et al.}~\cite{chang2015semantic}    & 0.096 \\
 Jiang \emph{et al.}~\cite{jiang2014zero}        & 0.101 \\
 Mazloom \emph{et al.}~\cite{MazloomICMR15}      & 0.119 \\
  \midrule
 \textbf{This paper}                             & \textbf{0.129} \\
 \bottomrule
 \end{tabular}
}
\end{table}
%

\section{Conclusion} \label{sec:conc}

This paper proposes TagBook, a new semantic video representation for video event detection.  
TagBook is based on freely available socially tagged videos, without the need for training any intermediate concept detectors. 
We introduce an algorithm that propagates tags to unlabeled videos from many socially tagged videos. 
The algorithm is inspired by image neighbor voting,
but is improved by refining the source set, i.e., removing existing noisy tags and generating new tags, before tag propagation. 
Experiments on the TRECVID 2013 and 2014 multimedia event detection datasets and the Columbia Consumer Video dataset show that TagBook outperforms the current state-of-the-art semantic video representations for both zero- and few-example video event detection.


\begin{IEEEbiography}[{\includegraphics[width=1in,height=1.25in,clip,keepaspectratio]{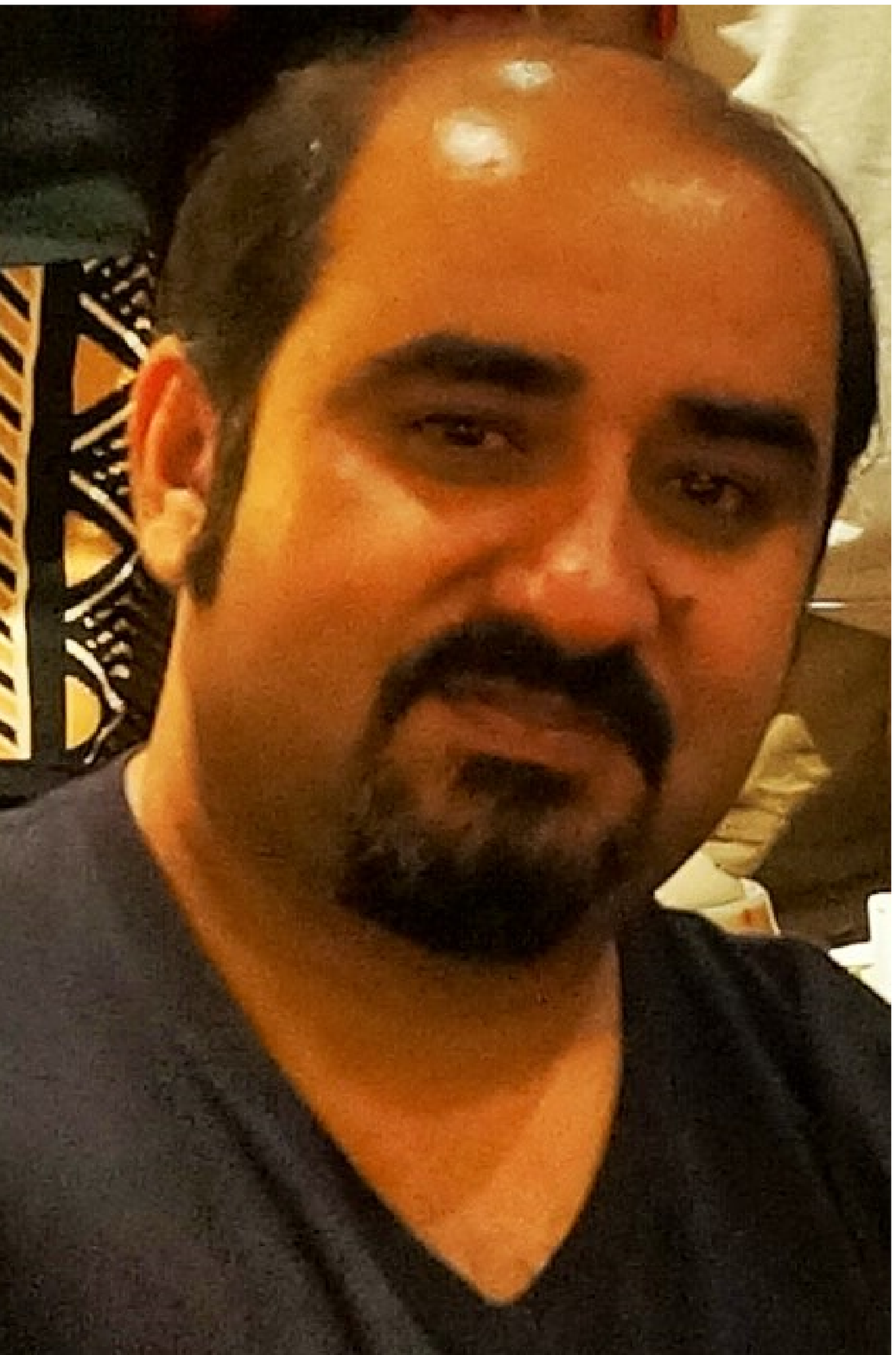}}]
{Masoud Mazloom} received the B.Sc. degree in computer engineering from Azad University, Tehran-South Campus in 2002, and the M.Sc. degree in computer science from Sharif University of Technology, Iran in 2005. After working as a lecturer at the Computer Engineering department in Shahid Chamran University, Ahvaz, Iran, he joined the Intelligent Systems Lab Amsterdam at the University of Amsterdam in 2011, where he received the Ph.D. degree in 2016. Currently he is a postdoctoral Researcher at the University of Amsterdam. His research interests focus on applying computer vision and machine learning algorithms for analyzing social media from marketing and business perspectives.
\end{IEEEbiography}

\begin{IEEEbiography}[{\includegraphics[width=1in,height=1.25in,clip,keepaspectratio]{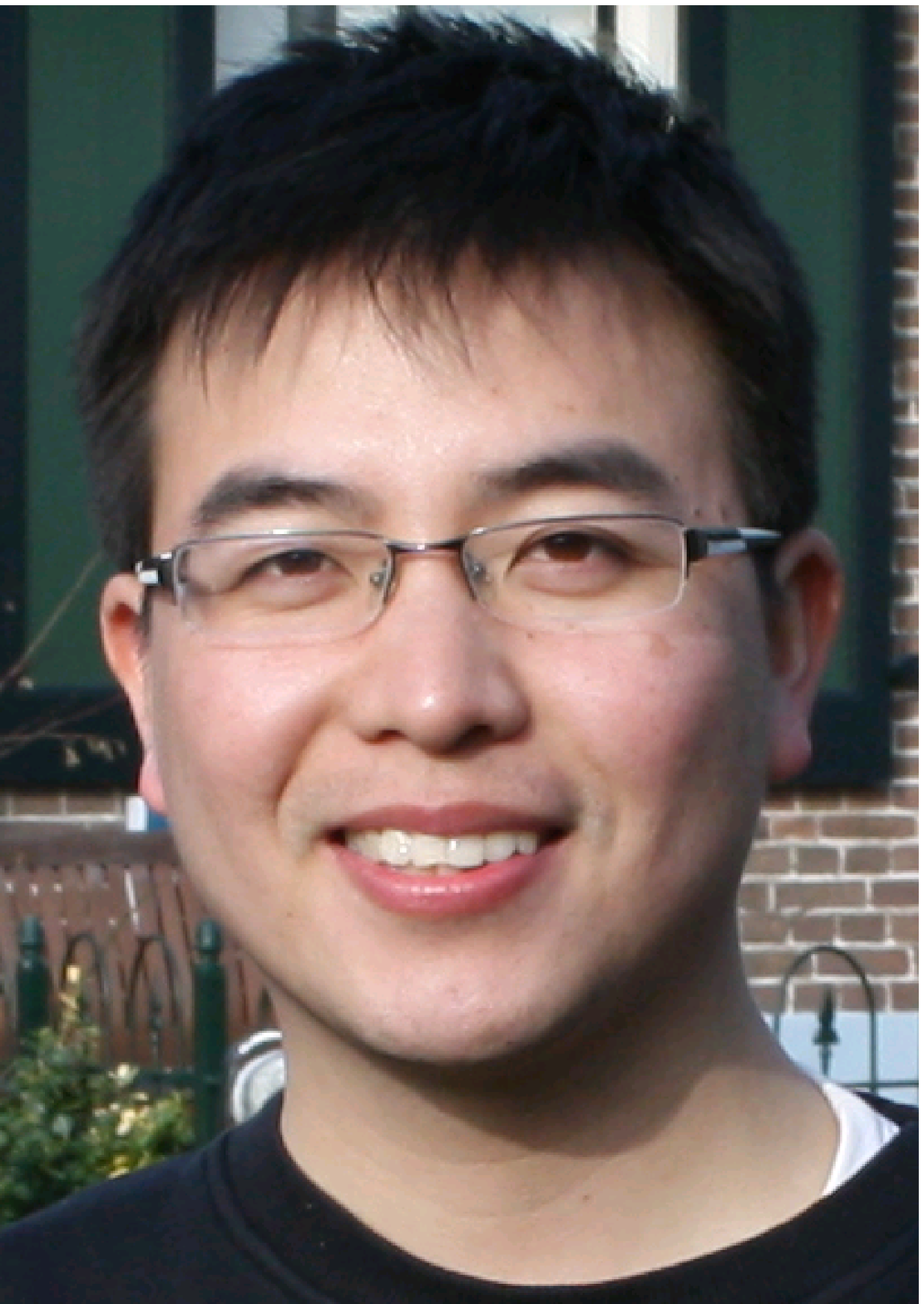}}]
{Xirong Li} received the B.S. and M.E. degrees from the Tsinghua University in 2005 and 2007, respectively,
and the Ph.D. degree from the University of Amsterdam in 2012, all in computer science.
He is currently an Assistant Professor in the Key Lab of Data Engineering and Knowledge Engineering,
Renmin University of China. His research focuses on image and video retrieval.
He has been awarded the ACM SIGMM Best PhD Thesis Award 2013, the IEEE TRANSACTIONS ON MULTIMEDIA Prize Paper Award 2012, the Best Paper Award of the ACM CIVR 2010, and PCM 2014 Outstanding Reviewer Award.
He was area chair of ICPR 2016 and publication co-chair of ICMR 2015.
\end{IEEEbiography}

\begin{IEEEbiography}[{\includegraphics[width=1in,height=1.25in,clip,keepaspectratio]{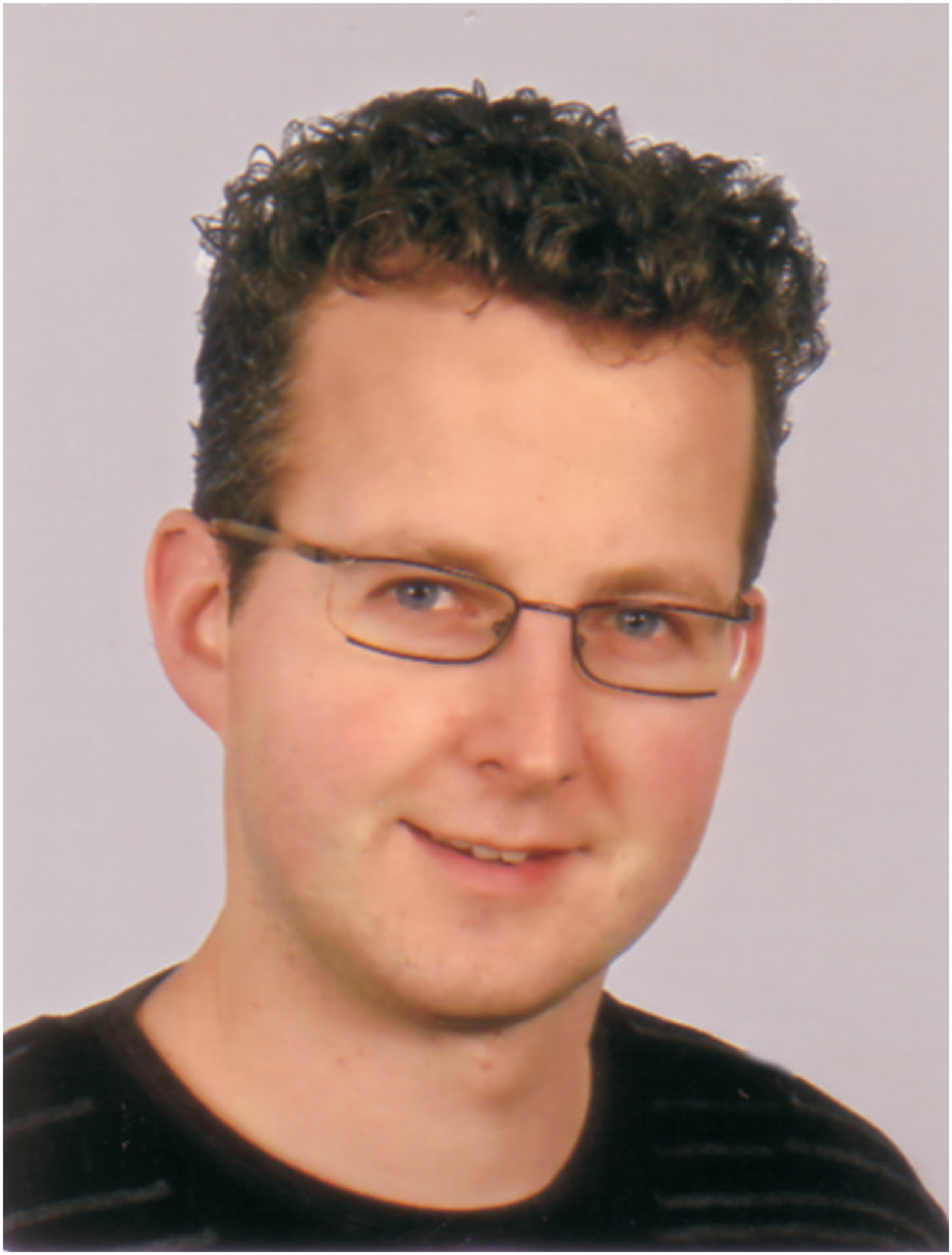}}]
{Cees G.M. Snoek} received the M.Sc. degree in business information systems (2000) and the Ph.D. degree in computer science (2005) both from the University of Amsterdam, The Netherlands. He is currently a director of the QUVA Lab, the joint research lab of Qualcomm and the University of Amsterdam on deep learning and computer vision. He is also a principal engineer at Qualcomm and an associate professor at the University of Amsterdam. He was previously a Visiting Scientist at Informedia, Carnegie Mellon University, USA (2003), Fulbright Junior Scholar at UC Berkeley’s Computer Vision Group (2010-2011), and head of R\&D at University spin-off Euvision Technologies before it was acquired by Qualcomm (2011-2014). His research interests focus on video and image recognition. He has published over 150 refereed book chapters, journal and conference papers, and serves on the program committee of the major conferences in multimedia and computer vision. Dr. Snoek is the lead researcher of the award-winning MediaMill Semantic Video Search Engine, which is the most consistent top performer in the yearly NIST TRECVID evaluations. He is general co-chair of ACM Multimedia 2016 in Amsterdam and program co-chair for ICMR 2017. He is a lecturer of post-doctoral courses given at international conferences and European summer schools. He is a senior member of IEEE and ACM. Dr. Snoek is member of the editorial boards for IEEE MultiMedia and ACM Transactions on Multimedia. Cees is recipient of an NWO Veni award (2008), a Fulbright Junior Scholarship (2010), an NWO Vidi award (2012), and the Netherlands Prize for ICT Research (2012). All for research excellence. Several of his Ph.D. students and Post-docs have won awards, including the IEEE Transactions on Multimedia Prize Paper Award (2012) the SIGMM Best Ph.D. Thesis Award (2013), the Best Paper Award of ACM Multimedia (2014) and an NWO Veni award (2015). Three of his former mentees serve as assistant professor.
\end{IEEEbiography}

\vfill


\begin{thebibliography}{10}
\providecommand{\url}[1]{#1}
\csname url@samestyle\endcsname
\providecommand{\newblock}{\relax}
\providecommand{\bibinfo}[2]{#2}
\providecommand{\BIBentrySTDinterwordspacing}{\spaceskip=0pt\relax}
\providecommand{\BIBentryALTinterwordstretchfactor}{4}
\providecommand{\BIBentryALTinterwordspacing}{\spaceskip=\fontdimen2\font plus
\BIBentryALTinterwordstretchfactor\fontdimen3\font minus
  \fontdimen4\font\relax}
\providecommand{\BIBforeignlanguage}[2]{{%
\expandafter\ifx\csname l@#1\endcsname\relax
\typeout{** WARNING: IEEEtran.bst: No hyphenation pattern has been}%
\typeout{** loaded for the language `#1'. Using the pattern for}%
\typeout{** the default language instead.}%
\else
\language=\csname l@#1\endcsname
\fi
#2}}
\providecommand{\BIBdecl}{\relax}
\BIBdecl

\bibitem{Ballan11}
L.~Ballan, M.~Bertini, A.~Del~Bimbo, L.~Seidenari, and G.~Serra, ``Event
  detection and recognition for semantic annotation of video,'' \emph{MTAP},
  vol.~51, 2011.

\bibitem{Lavee:understand}
G.~Lavee, E.~Rivlin, and M.~Rudzsky, ``Understanding video events: A survey of
  methods for automatic interpretation of semantic occurrences in videos,''
  \emph{TSMC}, vol.~39, no.~5, 2009.

\bibitem{Mubarak:ijmir}
Y.-G. Jiang, S.~Bhattacharya, S.-F. Chang, and M.~Shah, ``High-level event
  recognition in unconstrained videos,'' \emph{IJMIR}, vol.~2, no.~2, pp.
  73--101, 2013.

\bibitem{HAER00}
N.~Haering, R.~Qian, and I.~Sezan, ``A semantic event-detection approach and
  its application to detecting hunts in wildlife video,'' \emph{TCSVT},
  vol.~10, no.~6, pp. 857--868, 2000.

\bibitem{BONZ01}
A.~Bonzanini, R.~Leonardi, and P.~Migliorati, ``Event recognition in sport
  programs using low-level motion indices,'' in \emph{Proc. of ICME}, 2001.

\bibitem{LI01b}
B.~Li and M.~I. Sezan, ``Event detection and summarization in sports video,''
  in \emph{IEEE Workshop on Content-Based Access to Video and Image Libraries},
  2001.

\bibitem{BABA02}
N.~Babaguchi, Y.~Kawai, and T.~Kitahashi, ``Event based indexing of broadcasted
  sports video by intermodal collaboration,'' \emph{TMM}, vol.~4, no.~1, pp.
  68--75, 2002.

\bibitem{icmr11:CCV}
Y.-G. Jiang, G.~Ye, S.-F. Chang, D.~Ellis, and A.~C. Loui, ``Consumer video
  understanding: A benchmark database and an evaluation of human and machine
  performance,'' in \emph{Proc. of ICMR}, 2011.

\bibitem{MED13}
{NIST TRECVID Multimedia Event Detection (MED) Evaluation Track},
  \url{http://www.nist.gov/itl/iad/mig/med.cfm}.

\bibitem{Nakamasa:Tokyo}
N.~Inoue \emph{et~al.}, ``{TokyoTech+Canon at {TRECVID} 2011},'' in \emph{NIST
  TRECVID Workshop}, 2011.

\bibitem{CVPR2012BBN}
P.~Natarajan, S.~Wu, S.~N.~P. Vitaladevuni, X.~Zhuang, S.~Tsakalidis, U.~Park,
  R.~Prasad, and P.~Natarajan, ``Multimodal feature fusion for robust event
  detection in web videos,'' in \emph{Proc. of CVPR}, 2012.

\bibitem{OneataICCV2013}
D.~Oneata, J.~Verbeek, and C.~Schmid, ``Action and event recognition with
  fisher vectors on a compact feature set,'' in \emph{Proc. of ICCV}, 2013.

\bibitem{xu2015discriminative}
Z.~Xu, Y.~Yang, and A.~Hauptmann, ``A discriminative {CNN} video representation
  for event detection,'' in \emph{Proc. of CVPR}, 2015.

\bibitem{HabibianICM2014}
A.~Habibian, T.~Mensink, and C.~Snoek, ``{VideoStory}: A new multimedia
  embedding for few-example recognition and translation of events,'' in
  \emph{Proc. of ACM MM}, 2014.

\bibitem{shih-fu:2014}
J.~Chen, Y.~Cui, G.~Ye, D.~Liu, and S.-F. Chang, ``Event-driven semantic
  concept discovery by exploiting weakly tagged internet images,'' in
  \emph{Proc. of ICMR}, 2014.

\bibitem{WuCVPR2014}
S.~Wu, S.~Bondugula, F.~Luisier, X.~Zhuang, and P.~Natarajan, ``Zero-shot event
  detection using multi-modal fusion of weakly supervised concepts,'' in
  \emph{Proc. of CVPR}, 2014.

\bibitem{Jiang:Columbia}
Y.-G. Jiang, X.~Zeng, G.~Ye, S.~Bhattacharya, D.~Ellis, M.~Shah, and S.-F.
  Chang, ``{Columbia-UCF TRECVID2010} multimedia event detection: Combining
  multiple modalities, contextual concepts, and temporal matching,'' in
  \emph{NIST TRECVID Workshop}, 2010.

\bibitem{CVPR2012Kitware}
A.~Tamrakar, S.~Ali, Q.~Yu, J.~Liu, O.~Javed, A.~Divakaran, H.~Cheng, and H.~S.
  Sawhney, ``Evaluation of low-level features and their combinations for
  complex event detection in open source videos,'' in \emph{Proc. of CVPR},
  2012.

\bibitem{McCloskey:MVA13}
S.~Oh, S.~McCloskey, I.~Kim, A.~Vahdat, K.~Cannons, H.~Hajimirsadeghi, G.~Mori,
  A.~{Amitha Perera}, M.~Pandey, and J.~I. Corso, ``Multimedia event detection
  with multimodal feature fusion and temporal concept localization,''
  \emph{MVA}, vol.~25, no.~1, 2014.

\bibitem{habibian:MVA13}
G.~Myers, R.~Nallapati, J.~van Hout, S.~Pancoast, R.~Nevatia, C.~Sun,
  A.~Habibian, D.~Koelma, K.~van~de Sande, A.~Smeulders, and C.~Snoek,
  ``Evaluating multimedia features and fusion for example-based event
  detection,'' \emph{MVA}, vol.~25, no.~1, 2014.

\bibitem{super:icmr12}
Y.-G. Jiang, ``Super: towards real-time event recognition in internet videos,''
  in \emph{Proc. of ICMR}, 2012.

\bibitem{tmm2015-jiang-fast}
Y.-G. Jiang, Q.~Dai, T.~Mei, Y.~Rui, and S.-F. Chang, ``Super fast event
  recognition in internet videos,'' \emph{TMM}, vol.~17, no.~8, pp.
  1174–--1186, 2015.

\bibitem{dynamicpooling:iccv}
W.~Li, Q.~Yu, A.~Divakaran, and N.~Vasconcelos, ``Dynamic pooling for complex
  event recognition,'' in \emph{Proc. of ICCV}, 2013.

\bibitem{CVPR2014:dvmm}
K.-T. Lai, F.~Yu, M.-S. Chen, and S.-F. Chang, ``Video event detection by
  inferring temporal instance labels,'' in \emph{Proc. of CVPR}, 2014.

\bibitem{ECCV2014:dvmm}
K.-T. Lai, D.~Liu, M.-S. Chen, and S.-F. Chang, ``Recognizing complex events in
  videos by learning key static-dynamic evidences,'' in \emph{Proc. of ECCV},
  2014.

\bibitem{AXES}
D.~Oneata, M.~Douze, J.~Revaud, S.~Jochen, D.~Potapov, H.~Wang, Z.~Harchaoui,
  J.~Verbeek, C.~Schmid, R.~Aly, K.~Mcguiness, S.~Chen, N.~O'Connor,
  K.~Chatfield, O.~Parkhi, R.~Arandjelovic, A.~Zisserman, F.~Basura, and
  T.~Tuytelaars, ``{AXES at TRECVid 2012: KIS, INS, and MED},'' in \emph{NIST
  TRECVID Workshop}, 2012.

\bibitem{KrizhevskySH12}
A.~Krizhevsky, I.~Sutskever, and G.~E. Hinton, ``{ImageNet} classification with
  deep convolutional neural networks.'' in \emph{Proc. of NIPS}, 2012.

\bibitem{MazloomICMR15}
M.~Mazloom, A.~Habibian, D.~Liu, C.~Snoek, and S.-F. Chang, ``Encoding concept
  prototypes for video event detection and summarization,'' in \emph{Proc. of
  ICMR}, 2015.

\bibitem{ye2015}
G.~Ye, Y.~Li, H.~Xu, D.~Liu, and S.-F. Chang, ``{EventNet}: A large scale
  structured concept library for complex event detection in video,'' in
  \emph{Proc. of ACM MM}, 2015.

\bibitem{bmvc2015-nagel}
M.~Nagel, T.~Mensink, and C.~Snoek, ``Event fisher vectors: {R}obust encoding
  visual diversity of visual streams,'' in \emph{Proc. of BMVC}, 2015.

\bibitem{ma:acm13}
Z.~Ma, Y.~Yang, Z.~Xu, N.~Sebe, and A.~Hauptmann, ``We are not equally
  negative: {F}ine-grained labeling for multimedia event detection,'' in
  \emph{Proc. of ACM MM}, 2013.

\bibitem{Mazloom:acm13}
M.~Mazloom, A.~Habibian, and C.~Snoek, ``Querying for video events by semantic
  signatures from few examples,'' in \emph{Proc. of ACM MM}, 2013.

\bibitem{iccv2013-yang-few}
Y.~Yang, Z.~Ma, Z.~Xu, S.~Yan, and A.~Hauptmann, ``How related exemplars help
  complex event detection in web videos?'' in \emph{Proc. of ICCV}, 2013.

\bibitem{mm2015-chang-few}
X.~Chang, Y.-L. Yu, Y.~Yang, and A.~Hauptmann, ``Searching persuasively: Joint
  event detection and evidence recounting with limited supervision,'' in
  \emph{Proc. of ACM MM}, 2015.

\bibitem{habibian14icmr}
A.~Habibian, T.~Mensink, and C.~Snoek, ``Composite concept discovery for
  zero-shot video event detection,'' in \emph{Proc. of ICMR}, 2014.

\bibitem{jiang2014zero}
L.~Jiang, T.~Mitamura, S.-I. Yu, and A.~Hauptmann, ``Zero-example event search
  using multimodal pseudo relevance feedback,'' in \emph{Proc. of ICMR}, 2014.

\bibitem{chang2015semantic}
X.~Chang, Y.~Yang, A.~Hauptmann, E.~Xing, and Y.-L. Yu, ``Semantic concept
  discovery for large-scale zero-shot event detection,'' in \emph{Proc. of
  IJCAI}, 2015.

\bibitem{SMEA06}
A.~F. Smeaton, P.~Over, and W.~Kraaij, ``Evaluation campaigns and {TRECVid},''
  in \emph{Proc. of ACM MIR}, 2006.

\bibitem{NAPH06}
M.~R. Naphade, J.~R. Smith, J.~Te\v{s}i\'{c}, S.-F. Chang, W.~Hsu, L.~S.
  Kennedy, A.~G. Hauptmann, and J.~Curtis, ``Large-scale concept ontology for
  multimedia,'' \emph{{IEEE} MultiMedia}, vol.~13, no.~3, pp. 86--91, 2006.

\bibitem{ILSVRCarxiv14}
O.~Russakovsky, J.~Deng, H.~Su, J.~Krause, S.~Satheesh, S.~Ma, Z.~Huang,
  A.~Karpathy, A.~Khosla, M.~Bernstein, A.~Berg, and L.~Fei-Fei, ``{ImageNet
  Large Scale Visual Recognition Challenge},'' \emph{IJCV}, 2015, in press.

\bibitem{mtap15-svetlana}
S.~Kordumova, X.~Li, and C.~Snoek, ``Best practices for learning video concept
  detectors from social media examples,'' \emph{MTAP}, vol.~74, no.~4, pp.
  1291--1315, 2015.

\bibitem{Shahram:Multi}
S.~Ebadollahi, L.~Xie, S.-F. Chang, and J.~R. Smith, ``Visual event detection
  using multi-dimensional concept dynamics,'' in \emph{Proc. of ICME}, 2006.

\bibitem{Michele:SMV}
M.~Merler, B.~Huang, L.~Xie, G.~Hua, and A.~Natsev, ``Semantic model vectors
  for complex video event recognition,'' \emph{TMM}, vol.~14, no.~1, 2012.

\bibitem{HabibianCVIU14}
A.~Habibian and C.~Snoek, ``Recommendations for recognizing video events by
  concept vocabularies,'' \emph{CVIU}, vol. 124, pp. 110--122, 2014.

\bibitem{Izadinia:2012}
H.~Izadinia and M.~Shah, ``Recognizing complex events using large margin joint
  low-level event model,'' in \emph{Proc. of ECCV}, 2012.

\bibitem{Mazloom:TMM2014}
M.~Mazloom, E.~Gavves, and C.~Snoek, ``Conceptlets: Selective semantics for
  classifying video events,'' \emph{TMM}, vol.~16, no.~8, pp. 2214--2228, 2014.

\bibitem{GKAL11}
N.~Gkalelis, V.~Mezaris, and I.~Kompatsiaris, ``High-level event detection in
  video exploiting discriminant concepts,'' in \emph{Proc. of CBMI}, 2011.

\bibitem{Sun_2014_CVPR}
C.~Sun and R.~Nevatia, ``Discover: Discovering important segments for
  classification of video events and recounting,'' in \emph{Proc. of CVPR},
  2014.

\bibitem{cvpr14:trc}
S.~Bhattacharya, M.~Kalayeh, R.~Sukthankar, and M.~Shah, ``Recognition of
  complex events exploiting temporal dynamics between underlying concepts,'' in
  \emph{Proc. of CVPR}, 2014.

\bibitem{MA13}
Z.~Ma, ``From concepts to events: A progressive process for multimedia content
  analysis,'' Ph.D. dissertation, University of Trento, Trento, Italy, 2013.

\bibitem{2013trecvidover}
P.~Over, G.~Awad, M.~Michel, J.~Fiscus, G.~Sanders, W.~Kraaij, A.~F. Smeaton,
  and G.~Quéenot, ``{TRECVID} 2013 -- an overview of the goals, tasks, data,
  evaluation mechanisms and metrics,'' in \emph{NIST TRECVID Workshop}, 2013.

\bibitem{videoAtt:Haup}
Z.~Ma, Y.~Yang, Z.~Xu, N.~Sebe, S.~Yan, and A.~Hauptmann, ``Complex event
  detection via multi-source video attributes,'' in \emph{Proc. of CVPR}, 2013.

\bibitem{LiArxive15}
X.~Li, T.~Uricchio, L.~Ballan, M.~Bertini, C.~Snoek, and A.~{Del Bimbo},
  ``Socializing the semantic gap: A comparative survey on image tag assignment,
  refinement and retrieval,'' \emph{ACM Computing Surveys}, 2016, in press.

\bibitem{xirong:tmm09}
X.~Li, C.~Snoek, and M.~Worring, ``Learning social tag relevance by neighbor
  voting,'' \emph{TMM}, vol.~11, no.~7, pp. 1310--1322, 2009.

\bibitem{tagprop}
M.~Guillaumin, T.~Mensink, J.~Verbeek, and C.~Schmid, ``{TagProp}:
  Discriminative metric learning in nearest neighbor models for image
  auto-annotation,'' in \emph{Proc. of ICCV}, 2009.

\bibitem{Ballan:mm11}
L.~Ballan, M.~Bertini, A.~{Del Bimbo}, and G.~Serra, ``Enriching and localizing
  semantic tags in internet videos,'' in \emph{Proc. of ACM MM}, 2011.

\bibitem{stefan:sigir09}
S.~Siersdorfer, J.~S. Pedro, and M.~Sanderson, ``Automatic video tagging using
  content redundancy,'' in \emph{Proc. of SIGIR}, 2009.

\bibitem{Mazloom:ICMR14}
M.~Mazloom, X.~Li, and C.~Snoek, ``Few-example video event retrieval using tag
  propagation,'' in \emph{Proc. of ICMR}, 2014.

\bibitem{pegasos}
S.~Shalev-Shwartz, Y.~Singer, N.~Srebro, and A.~Cotter, ``Pegasos: primal
  estimated sub-gradient solver for svm,'' \emph{Math. Program.}, vol. 127,
  no.~1, 2011.

\bibitem{pami2012-efm}
A.~Vedaldi and A.~Zisserman, ``Efficient additive kernels via explicit feature
  maps,'' \emph{IEEE Trans. Pattern Anal. Mach. Intell.}, vol.~34, no.~3, pp.
  480–--492, 2012.

\bibitem{med2014:cmu}
S.-I. Yu, L.~Jiang, Z.~Xu, Z.~Lan, S.~Xu, X.~Chang, X.~Li, Z.~Mao, C.~Gan,
  Y.~Miao, X.~Du, Y.~Cai, L.~Martin, N.~Wolfe, A.~Kumar, H.~Li, M.~Lin, Z.~Ma,
  Y.~Yang, D.~Meng, S.~Shan, P.~Sahin, S.~Burger, F.~Metze, R.~Singh, B.~Raj,
  T.~Mitamura, R.~Stern, and A.~Hauptmann, ``Informedia@ {TRECVID} 2014 {MED}
  and {MER},'' in \emph{NIST TRECVID Workshop}, 2014.

\bibitem{med2015:uva}
C.~Snoek, S.~Cappallo, D.~Fontijne, D.~Julian, D.~Koelma, P.~Mettes, K.~van~de
  Sande, A.~Sarah, H.~Stokman, and R.~Towal, ``{Qualcomm Research and
  University of Amsterdam at {TRECVID} 2015: {R}ecognizing Concepts, Objects,
  and Events in Video},'' in \emph{NIST TRECVID Workshop}, 2015.

\bibitem{mm2015jiang-cmu}
L.~Jiang, S.-I. Yu, D.~Meng, Y.~Yang, T.~Mitamura, and A.~Hauptmann, ``Fast and
  accurate content-based semantic search in 100m {Internet} videos,'' in
  \emph{Proc. of ACM MM}, 2015.

\end{thebibliography}
\end{document}